%% file: main.tex
\documentclass{article}
\usepackage{iclr2026_conference,times}

\usepackage{hyperref}
\usepackage{url}
\usepackage{graphicx}
\usepackage{subcaption}
\usepackage{xcolor}
\usepackage{xspace}

\usepackage{listings}
\definecolor{codegreen}{rgb}{0.25,0.49,0.37}
\definecolor{codepurple}{rgb}{0.58,0,0.82}
\definecolor{codered}{rgb}{0.73,0.13,0.13}
\definecolor{codegray}{rgb}{0.5,0.5,0.5}
\definecolor{backgray}{rgb}{0.97,0.97,0.97}
\title{Advancing Earth Observation Through \\ Machine Learning: A TorchGeo Tutorial}

\author{
Caleb Robinson$^{1}$\thanks{\texttt{caleb.robinson@microsoft.com}} \hspace{0.9em}
Nils Lehmann$^{2}$ \hspace{0.9em}
Adam J. Stewart$^{2}$ \hspace{0.9em}
Burak Ekim$^{3}$ \\
\bf Heng Fang$^{4}$ \hspace{0.9em}
Isaac A. Corley$^{5}$ \hspace{0.9em}
Maur\'{\i}cio Cordeiro$^{6}$
\\[1.5em]
\small \textsuperscript{1}Microsoft AI for Good Research Lab \\
\small \textsuperscript{2}Chair of Data Science in Earth Observation, Technical University of Munich \\
\small \textsuperscript{3}Chair of Earth Observation, University of the Bundeswehr Munich \\
\small \textsuperscript{4}Department of Robotics, Perception and Learning, KTH Royal Institute of Technology \\
\small \textsuperscript{5}Wherobots \\
\small \textsuperscript{6}National Water and Sanitation Agency, Brazil, Federal District, Brasilia
}

\iclrfinalcopy %

\begin{document}

\maketitle

\begin{abstract}
Earth observation machine learning pipelines differ fundamentally from standard computer vision workflows. Imagery is typically delivered as large, georeferenced scenes, labels may be raster masks or vector geometries in distinct coordinate reference systems, and both training and evaluation often require spatially aware sampling and splitting strategies. TorchGeo is a PyTorch-based domain library that provides datasets, samplers, transforms and pre-trained models with the goal of making it easy to use geospatial data in machine learning pipelines. In this paper, we introduce a tutorial that demonstrates 1.) the core TorchGeo abstractions through code examples, and 2.) an end-to-end case study on multispectral water segmentation from Sentinel-2 imagery using the Earth Surface Water dataset. This demonstrates how to train a semantic segmentation model using TorchGeo datasets, apply the model to a Sentinel-2 scene over Rio de Janeiro, Brazil, and save the resulting predictions as a GeoTIFF for further geospatial analysis. The tutorial code itself is distributed as two Python notebooks:  \href{https://torchgeo.readthedocs.io/en/stable/tutorials/torchgeo.html}{Introduction to TorchGeo} and 
\href{https://torchgeo.readthedocs.io/en/stable/tutorials/earth_surface_water.html}{Earth Surface Water}.

\end{abstract}

\section{Motivation and tutorial scope}
Earth observation (EO) has become a central input to scientific monitoring and decision making, and modern machine learning (ML) methods are increasingly used to extract structure from satellite imagery at scale~\citep{reichstein2019deep}. However, satellite data is a distinct modality whose data distributions, metadata, and evaluation requirements differ in consequential ways from standard natural-image workflows \citep{rolf2024position} --- using satellite imagery is \emph{not} simply ``computer vision with larger images''.

In typical EO pipelines, imagery is delivered as large, georeferenced scenes (often multi-band), while labels may be distributed separately as raster masks or vector geometries in different coordinate reference systems (CRSs) and at different spatial resolutions. Training and evaluation therefore require operations that are not standard in vision workflows but central in EO: reprojecting and resampling layers into a common grid, sampling geographically aligned chips on-the-fly from scenes that are too large to fit into memory, constructing spatially separated train/validation/test splits to reduce label leakage, and exporting predictions back to georeferenced products for mapping and downstream analyses. Even when using standard model architectures (e.g., a semantic segmentation network), substantial domain specific work is still needed to correctly align inputs and targets and ensure that the experimental protocol is spatially meaningful and reproducible.

TorchGeo was developed to address this friction by providing a PyTorch~\citep{paszke2019pytorch} domain library, similar to torchvision~\citep{TorchVision_maintainers_and_contributors_TorchVision_PyTorch_s_Computer_2016}, with geospatial datasets, geographic samplers, transforms, and pretrained-model interfaces designed to handle geospatial metadata~\citep{stewart2025torchgeo}. The goal of this tutorial is to introduce the main abstractions used in TorchGeo to readers, and show how these fit together in an end-to-end applied workflow in two executable Python notebooks. First, the \emph{Introduction to TorchGeo} notebook demonstrates the minimal set of primitives needed for most EO learning problems: composing partially overlapping layers (e.g., mosaics and label rasters), indexing datasets by spatial windows, and using geographic samplers to generate training chips without pre-tiling. Second, the \emph{Earth Surface Water} notebook shows an example of multispectral semantic segmentation with Sentinel-2 imagery and binary water masks, including practical preprocessing (reflectance scaling, normalization, and adding spectral indices as input channels), model adaptation to \(C>3\) channels, and spatial evaluation. Importantly, it also shows how to run gridded inference with overlapping patches using a trained model with a Sentinel-2 scene over Rio de Janeiro and save the predictions as a single GeoTIFF.

\section{Introduction to TorchGeo}
We begin with a short sequence of code snippets (with accompanying explanations in the tutorial video/notebook) adapted from the TorchGeo ``Introduction to TorchGeo'' tutorial \citep{torchgeo_tutorial_intro}. The goal is to make the key abstractions concrete in a few lines, rather than to present an survey of the entire API. Link to the tutorial is \href{https://torchgeo.readthedocs.io/en/stable/tutorials/torchgeo.html}{here}.

\paragraph{Composable datasets: union and intersection}
A common EO workflow requires combining multiple partially overlapping rasters (e.g., multi-sensor imagery) and sampling only where both imagery and labels exist. TorchGeo exposes this directly through intersection ``$\&$'', and union ``$|$'', set-like composition operators:
\begin{lstlisting}
# landsat7, landsat8 and cdl are all different torchgeo RasterDatasets
landsat = landsat7 | landsat8 
dataset = landsat & cdl
\end{lstlisting}
The union operator constructs a virtual mosaic of all available imagery tiles, while the intersection operator restricts sampling to regions where inputs and targets overlap. Importantly, this composition is lazy: TorchGeo does not pre-mosaic or pre-reproject entire scenes, but performs windowed reads and alignment on demand.

\paragraph{Spatiotemporal indexing into large rasters}
Unlike curated image datasets, geospatial layers are indexed by location (and often time). TorchGeo supports slicing by projected coordinates and timestamps to retrieve an aligned chip:
\begin{lstlisting}
size = 256
xmin = 925000  # geographic coordinate
xmax = xmin + size * 30  # 30m pixel resolution
ymin = 4470000
ymax = ymin + size * 30

sample = dataset[xmin:xmax, ymin:ymax]  # slicing happens in geospatial coordinates
\end{lstlisting}
In the tutorial, this example is used to emphasize that the user can request a small, pixel-aligned patch from multiple large rasters without manual preprocessing, and the library ensures consistent CRS and spatial resolution between layers at read time. Practically, users rely on geographic samplers, described in the next paragraphs, to generate these slices on demand for their use case (i.e. model training or inference).

\paragraph{Geographic samplers and PyTorch data loaders}
Finally, we show how geographic sampling integrates with standard PyTorch training loops. Random sampling is typically preferred for training, while gridded sampling is preferred for evaluation and inference:
\begin{lstlisting}
train_sampler = RandomGeoSampler(dataset, size=size, length=1000)
test_sampler = GridGeoSampler(dataset, size=size, stride=size)

# torchgeo datasets and samplers work with the standard PyTorch DataLoader
train_dataloader = DataLoader(
    dataset, batch_size=128, sampler=train_sampler, collate_fn=stack_samples
)
test_dataloader = DataLoader(
    dataset, batch_size=128, sampler=test_sampler, collate_fn=stack_samples
)
\end{lstlisting}
This design keeps the surrounding modeling code close to ``standard'' PyTorch: once chips are produced, the remaining steps (model definition, loss, optimizer, logging) closely mirror conventional computer vision practice. Lazy loading via \texttt{RandomGeoSampler} bypasses the storage-heavy `pre-tiling' stage by performing windowed reads of only the required pixels for each batch. Today, TorchGeo contains more than 140 datasets and more than 60 diverse datamodules that can be readily used with PyTorch Lightning~\citep{Falcon_PyTorch_Lightning_2019}.

\section{Case study: water segmentation on Earth Surface Water}
The second half of the tutorial develops an end-to-end notebook based on the TorchGeo ``Earth Water Surface'' case study \citep{torchgeo_tutorial_earth_surface_water}. The Earth Surface Water dataset \citep{luo2021applicable,xin_luo_2021_5205674} contains Sentinel-2 image patches and corresponding binary water masks drawn from diverse geographic regions. The goal of the tutorial is to demonstrate a real-world EO segmentation workflow, with emphasis on steps that frequently cause failures in ad hoc pipelines: multispectral handling, CRS alignment across globally distributed tiles, and sampler configuration. Link to the tutorial \href{https://torchgeo.readthedocs.io/en/stable/tutorials/earth_surface_water.html}{here}.

\paragraph{Creating paired raster datasets and handling CRS globally.}
The tutorial constructs separate \texttt{RasterDataset} objects for imagery and masks, applies a reflectance scaling transform to imagery, marks mask rasters as non-image targets, and then pairs inputs and labels with the intersection operator. Because patches are distributed worldwide (often in different UTM zones), the notebook specifies a global CRS (World Mercator, EPSG:3395) so that all samples are consistently aligned during sampling and loading.

\paragraph{Sampling and batching.}
Training and validation chips are sampled with \texttt{RandomGeoSampler} using a fixed chip size (e.g., \(512\times512\)~px). Sampling is performed in geographic space --- mitigating the effects of spatial autocorrelation to improve model generalization --- while yielding standard tensor batches suitable for GPU training. The notebook includes batch visualization utilities to sanity-check image/mask alignment and class balance before training.

\paragraph{Multispectral transforms and spectral indices.}
The case study demonstrates a realistic multispectral preprocessing chain. It computes dataset-specific mean and standard deviation statistics over the training imagery and then appends spectral indices --- two NDWI variants (using different NIR bands) and NDVI --- before normalization. Because indices are appended as additional channels, the notebook pads the normalization statistics so that normalization is applied to the original sensor bands while leaving the appended indices unchanged. We find that this pattern is a useful template for many EO tasks where learned models benefit from physically meaningful indices but training stability benefits from normalization.

\paragraph{Model choice and adapting an RGB architecture to $C>3$ channels.}
For the segmentation model, the notebook uses DeepLabV3~\citep{chen2017rethinking} with a ResNet-50 backbone~\citep{he2016deep} from torchvision, configured for two output classes and trained from scratch (pretrained weights are omitted because they target RGB natural images). The critical EO-specific adaptation to the model initialization is adapting the first convolutional layer to accept the transformed input dimensionality (e.g., 6 spectral channels and 3 computed band indices). The notebook re-initializes the backbone's first convolutional layer to accept the multispectral channel count while preserving the output feature dimension.

\paragraph{Training and evaluation.}
The tutorial includes a compact training loop (run on a GPU, if available) that uses a standard segmentation loss, and evaluates predictions with an IoU/Jaccard metric. We train the model from scratch for 10 epochs with 130 samples per epoch, using an AdamW optimizer with a learning rate of 0.0001, and achieves an overall accuracy of 0.977 and IoU of 0.824 on the val set.

\input{figures/earth_water_prediction}

\paragraph{Inference over Rio de Janeiro}
The final, critical, component of the tutorial involves running the trained model on a Sentinel-2 scene over Rio de Janeiro, Brazil captured on 02/01/2026 from the Microsoft Planetary Computer\footnote{The scene ID is \texttt{S2C\_MSIL2A\_20260201T130241\_R095\_T23KPQ\_20260201T153209} although any Sentinel-2 scene could be used.}. Here, we are running the model on a grid of overlapping image patches and saving the resulting predictions as a GeoTIFF (i.e. that is pixel aligned with the Sentinel-2 input). A crop of the scene and resulting prediction is shown in Figure \ref{fig:earth_surface_pred}. This step allows users to actually use the model in cases that they care about and explore its limits past looking at test set metrics (under what conditions does it fail? how sharp are the predictions in along coastlines and through rivers? what is the smallest water feature that the model can detect?).

\clearpage

\bibliography{iclr2026_conference}
\bibliographystyle{iclr2026_conference}

\end{document}

%% file: figures/earth_water_prediction.tex
\begin{figure}[htbp]
    \centering
    \includegraphics[width=0.8\textwidth]{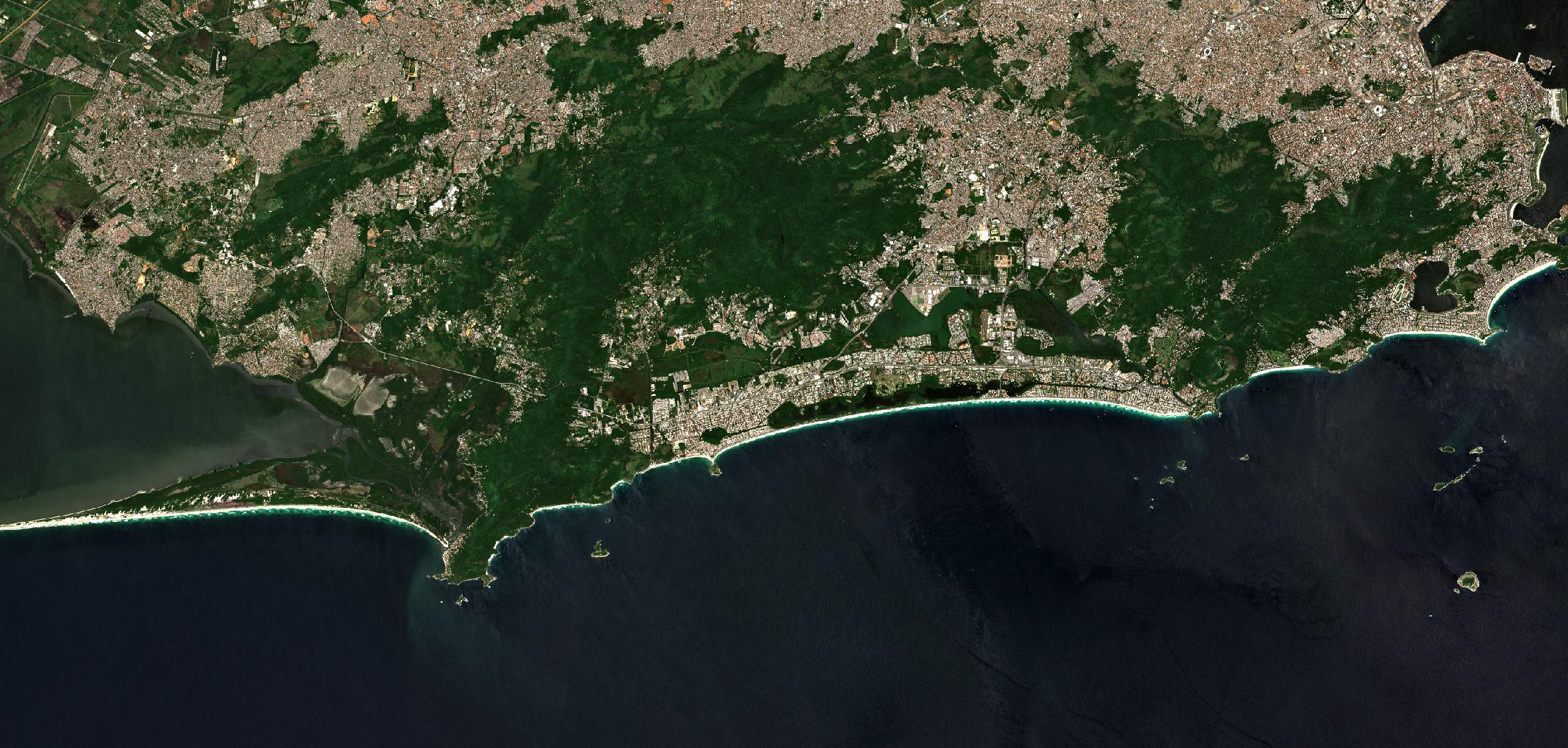}
    \includegraphics[width=0.8\textwidth]{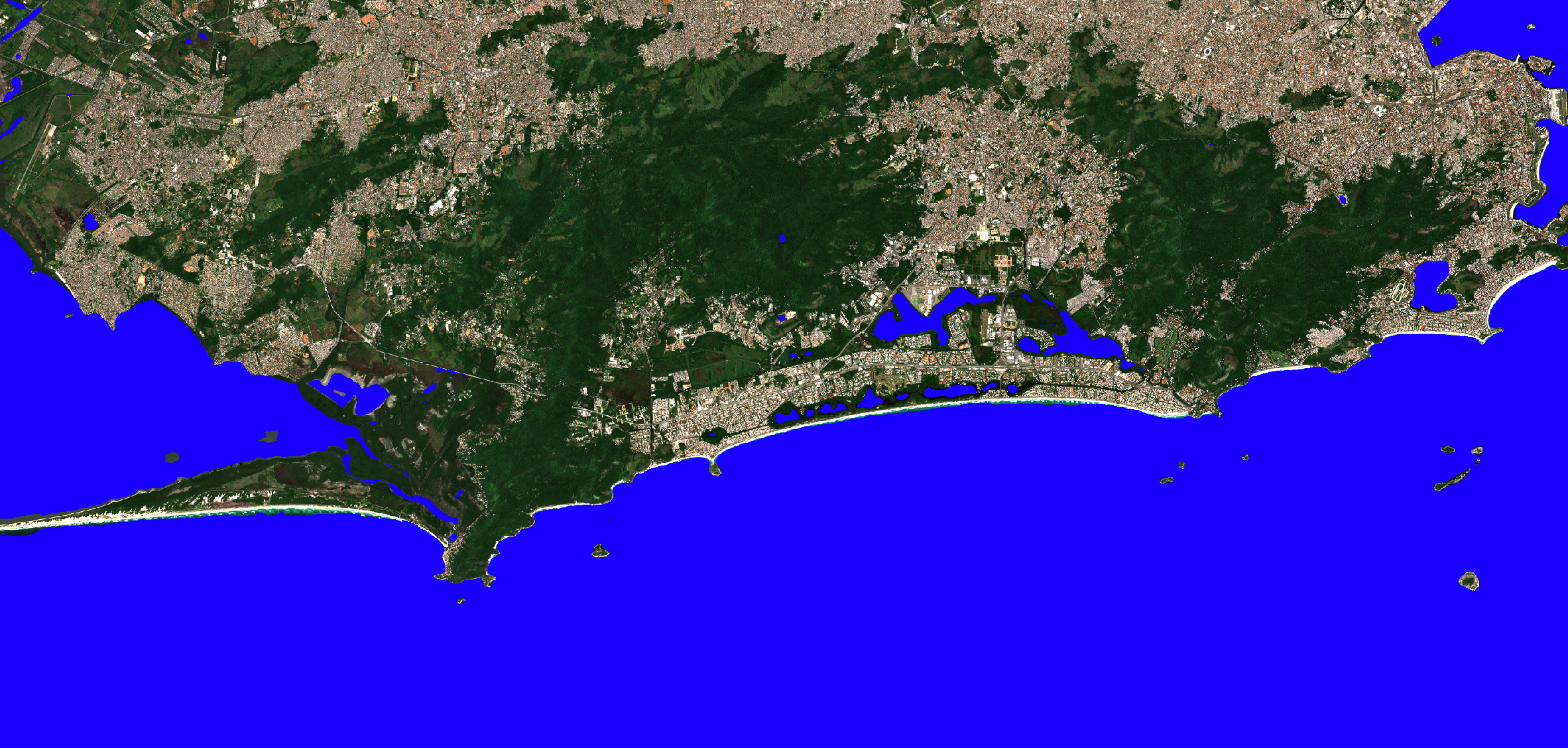}
    \caption{Earth Surface Water trained model predictions (blue) on a Sentinel-2 scene captured over Rio de Janeiro, Brazil.}
    \label{fig:earth_surface_pred}
\end{figure}